\documentclass{article}

\usepackage[final]{nips_2018} %

\usepackage[utf8]{inputenc} %
\usepackage[T1]{fontenc}    %

\usepackage{tipa}

\usepackage{hyperref}       %
\usepackage{url}            %
\usepackage{booktabs}       %
\usepackage{amsfonts}       %
\usepackage{nicefrac}       %
\usepackage{microtype}      %
\usepackage{wrapfig}
\usepackage{textcomp}
\usepackage{gensymb}        %
\usepackage{multirow}

\usepackage{framed}

\usepackage{algorithm}
\usepackage{algpseudocode}

\usepackage{amsmath,amssymb}
\usepackage{mathtools}
\usepackage{enumitem}
\usepackage[noabbrev,capitalize]{cleveref}
\usepackage{icomma}
\usepackage{xspace}
\usepackage{color}
\usepackage{wrapfig}
\usepackage{subcaption}

\newcommand{\be}{\begin{equation}}
\newcommand{\ee}{\end{equation}}
\newcommand{\bea}{\begin{eqnarray}}
\newcommand{\eea}{\end{eqnarray}}
\newcommand{\beaa}{\begin{eqnarray*}}
\newcommand{\eeaa}{\end{eqnarray*}}

\DeclareMathAlphabet{\mathpzc}{OT1}{pzc}{m}{n}

\title{Composing RNNs and FSTs for Small Data: \\Recovering Missing Characters in Old Hawaiian Text}

\author{
  Oiwi Parker Jones\thanks{~Authors contributed equally.}~~\textsuperscript{1,2}
  \textnormal{and}
  Brendan Shillingford\footnotemark[1]~~\textsuperscript{2,3}
\\
  Jesus College\textsuperscript{1},
  University of Oxford\textsuperscript{2}, 
  Google DeepMind\textsuperscript{3} \\
  \texttt{oiwi.parkerjones@jesus.ox.ac.uk}, \\
  \texttt{brendan.shillingford@cs.ox.ac.uk}
}

\newcommand{\orthog}[1]{{$\langle${#1}$\rangle$}}

\newcommand{\bs}[1]{}

\usepackage{xifthen}

\newcommand{\fstngram}[3]{\textsc{FST-}{#1}\textsc{-{#2}gram{\ifthenelse{\equal{#3}{}}{}{-#3}}\xspace}}

\newcommand{\fstrnn}{\textsc{FST-RNNLM}\xspace}

\newcommand{\correct}[1]{\textcolor{green}{\textbf{#1}}}
\newcommand{\correctmarker}{\correct{green and bold}}

\newcommand{\wrong}[1]{\textcolor{red}{\underline{#1}}}
\newcommand{\wrongmarker}{\wrong{red underline}}
\newcommand{\missing}[1]{\textcolor{blue}{\textit{#1}}}
\newcommand{\missingmarker}{\missing{blue and italic}}

\newcommand{\samp}[1]{
\par
\noindent
\hangindent=.5cm
\textit{#1}:}
\newcommand{\samprule}{
\noindent
{\hfill \rule{.3\linewidth}{0.2pt} \hfill}
}

\begin{document}

\maketitle

\begin{abstract}

In contrast to the older writing system of the 19th century, modern Hawaiian orthography employs characters for long vowels and glottal stops. 
These extra characters account for about one-third of the phonemes in Hawaiian, so including them makes a big difference to reading comprehension and pronunciation. 
However, transliterating between older and newer texts is a laborious task when performed manually. 
We introduce two related methods to help solve this transliteration problem automatically, given that there were not enough data to train an end-to-end deep learning model. 
One method is implemented, end-to-end, using finite state transducers (FSTs). 
The other is a hybrid deep learning approach which approximately composes an FST with a recurrent neural network (RNN). %
We find that the hybrid approach outperforms the end-to-end FST by partitioning the original problem into one part that can be modelled by hand, using an FST, and into another part, which is easily solved by an RNN trained on the available data.

\end{abstract}

\section{Introduction}
\label{sec:intro}
From 1834 to 1948, more than 125,000 newspaper pages were published in the Hawaiian language \citep{nogelmeier2010}. 
Yet by 1981, many expected this once flourishing language to die \citep{benton1981}. 
Hawaiian has since defied expectations and experienced the beginnings of a remarkable recovery \citep{warner2001, wilson+2001}. 
However much of the literary inheritance that is contained in the newspapers has become difficult for modern Hawaiians to read, since the newspapers were written in an orthography that failed to represent about one-third of the language's phonemes. 
This orthography, which we will refer to as the \emph{missionary orthography}, excluded Hawaiian phonemes that did not have equivalents in American English \citep[see][]{schutz1994}, namely long vowels \textipa{/i: e: a: o: u:/} and glottal stop \textipa{/P/}. 
By contrast, the \emph{modern Hawaiian orthography}, an innovation of Pukui and Elbert's Hawaiian dictionary \citep{Pukui+1957}, presents a nearly perfect, one-to-one mapping between graphemes and phonemes. 
The process of manual transliteration from missionary to modern Hawaiian orthography is extremely labor intensive. 
Yet the cultural benefits are so great that hundreds of pages of newspaper-serials have already been transliterated by hand, such as Nogelmeier's new edition of the epic tale of \textit{Hi`iakaikapoliopele}, the volcano goddess's sister \citep{hooulumahiehie2007}. 
Critically important as such efforts are to the continued revitalization of this endangered language, they are still only a small sample of the material that could be made available to a modern Hawaiian audience. 

In this paper, we propose to automate, or semi-automate, the transliteration of old Hawaiian texts into the modern orthography. 
Following a brief review of related work (\cref{sec:related-work}), we begin by describing a dataset of modern Hawaiian (\cref{sec:data-preproc}). %
In \cref{sec:models},
we present two methods for recovering missing graphemes (and hence phonemes) from the missionary orthography. 
The first composes a series of weighted FSTs; the second approximately composes an FST with a recurrent neural network language model (RNNLM) using a beam search procedure.
Both approaches require only modern Hawaiian texts for training, which are much more plentiful than parallel corpora. 
\cref{sec:results} reports the results of our transliteration experiments using a simulated parallel corpus, as well as two 19th century newspaper articles for which we also have modern Hawaiian transcriptions.
Being based on FSTs, both approaches are modular and extensible.
We observe useful and promising results for both of our methods, with the best results obtained by the hybrid FST-RNNLM. 
These results showcase the strength of combining established hand-engineering methods with deep learning in a smaller data regime, with practical applications for an endangered language. 

\section{Related work}
\label{sec:related-work}
Many of the themes that we address relate to existing literature. 
For example, \citet{hajic+2000} and \citet{scannell2014} have written on machine translation (MT) for closely related languages and on multilingual text normalization, respectively. 
Though language-relatedness makes MT easier \citep{kolovratnik+2009}, %
state-of-the-art techniques such as neural machine translation (NMT) 
have not performed well for languages with little data \citep{ostling+2017}. 
So while the Hawaiian transliteration problem could be cast as an instance of MT or of NMT, we chose to sidestep the scarcity of parallel data by not considering such approaches.

Hybrid approaches, which combine expert knowledge for well-understood structures with deep learning for data-plentiful subproblems, offer rich opportunities for data-efficient modelling.
In particular, prior work has explored ways to combine FSTs and RNNs. 
For example, \citet{sproat+2016} used an FST to restrict the search space when decoding from an RNN; \citet{rastogi+2016} incorporated RNN information into an FST.
However, these approaches differ from the approximate FST-to-RNN composition algorithm that we introduce here (in \cref{sec:hybrid-models}).

\section{Data}
\label{sec:data-preproc}

\subsection{Phonemes \& the modern orthography}
\label{sec:phonemes}

Ignoring case, there is a neat mapping between the modern Hawaiian orthography and the Hawaiian phonemic inventory. 
The phonemic inventory contains eight consonants \textipa{/h k l m n p v P/} and ten vowels, of which five are short \textipa{/a e i o u/} and five are long \textipa{/a: e: i: o: u:/} \citep{parkerjones2018}. 

The consonants map onto the orthographic symbols \orthog{H h K k L l M m N n P p W w `}, where we give the upper- and lower-case variants in adjacent pairs: \orthog{H h} for \textipa{/h/}, \orthog{K k} for \textipa{/k/}, \dots, \orthog{W w} for \textipa{/v/}. An exception, the symbol \orthog{`} has only one variant which maps to \textipa{/P/}. 
The vowels map onto the symbols: \orthog{A a E e I i O o U u \={A} \={a} \={E} \={e} \={I} \={\i} \={O} \={o} \={U} \={u}}. 
Note that vowel length is denoted by the absence or presence of a macron (e.g.~\orthog{A} and \orthog{a} map onto short \textipa{/a/} and \orthog{\={A}} and \orthog{\={a}} map onto long \textipa{/a:/}).  

The Hawaiian conventions for capitalization, numbering, and punctuation are analogous to those in English, except again there is no upper-case variant of \orthog{`}, so the following vowel is capitalized instead (e.g.~\textit{`Okakopa} `October').
In foreign words, such as \textit{kolorofolorokalapona} `chlorofluorocarbon', one can find the additional consonants: \orthog{B b C c D d F f G g J j Q q R r S s T t V v X x Y y Z z}.

\subsection{Modern \& missionary orthographies}
\label{sec:orthographies}

The primary difference between the missionary and modern Hawaiian orthographies is that the missionary orthography does not encode long vowels or the glottal stop. 
For example, the following Hawaiian phrases were recorded by a 19th-century German traveller in the missionary orthography: \textit{Ua oia au}, \textit{E ue ae oe ia Ii}, \textit{E ao ae oe ia ia} \cite[p.~7]{chamisso1837}. 
In the modern orthography these become: \textit{Ua `\={o} `ia au} `I am speared', \textit{E u\={e} a`e `oe i\={a} `\={I}`\={\i}} `You must weep for `\={I}`\={\i} (a person)', and \textit{E a`o a`e `oe i\={a} ia} `You teach him' \cite[p.~3]{elbert+1979}. %

We can convert text in the modern Hawaiian orthography \emph{backward} chronologically to an approximate missionary orthography by mapping each glottal stop \orthog{`} to the empty string $\epsilon$, and each long vowel, e.g.~\orthog{\={a} \={e} \={\i} \={o} \={u}}, to its corresponding short vowel, \orthog{a e i o u}. 
As a first approximation, we may treat mappings from the modern-to-missionary orthographies as unambiguously many-to-one; thus there is information loss. 
We will return to secondary differences between the orthographies in \cref{sec:conclusions}. 
To illustrate, the following four words in the modern orthography all map to the same missionary string \textit{aa}: \textit{a`a} (root), \textit{`a`a} (brave), \textit{`a`\={a}} (crumbly lava rock), and \textit{`\={a}`\={a}} (stutter).

The \emph{forward} mapping from missionary-to-modern orthographies is one-to-many. 
Thus the missionary string \textit{aa} could map to \textit{a`a}, \textit{`a`a}, \textit{`a`\={a}}, or \textit{`\={a}`\={a}}. 
The \emph{transliteration problem} we address here seeks to discover how we can use context to recover the information not present in the missionary orthography that modern Hawaiian orthography retains.

\subsection{Data sources}
\label{sec:data-preproc:corpus}

We draw on three sources for modern Hawaiian text: the main text of \textit{Hi`iakaikapoliopele} \citep{hooulumahiehie2007}, 160 short texts from \textit{Ulukau: The Hawaiian Electronic Library}, and the full Hawaiian Wikipedia (see \cref{fig:dataset}).\footnote{\textit{Ulukau: The Hawaiian Electronic Library}: \url{http://ulukau.org/}, Hawaiian Wikipedia: \url{https://haw.wikipedia.org/}. Both accessed 19 May 2018.}

For evaluation, we simulate a missionary-era version of the modern texts using the backward mapping described above. 
In addition, we evaluated our models on a couple of 19th century newspaper samples for which we have parallel missionary-era and modern text. 
Both simulated and real parallel corpora will be described in \cref{sec:results}.

\begin{table}
  \begin{center}
  \begin{tabular}{l|r|r}
    \toprule
    Source & Chars & Words\\
    \midrule
    Ulukau%
    (160 texts) & 6,518,451 & 1,334,451 \\
    Hi`iakaikapoliopele & 1,272,935 & 259,947 \\
    Wikipedia%
    & 577,794 & 10,221 \\
    \midrule
    \textit{Total} & {8,369,180} & {1,604,619} \\
    \bottomrule
  \end{tabular}
  \vspace{.5em}
  \caption{Modern data sources and their sizes.}
  \vspace{-1em}
  \label{fig:dataset}
  \end{center}
\end{table}

\section{Models}
\label{sec:models}
We can frame the task of transliterating from missionary-to-modern Hawaiian orthographies as a sequence transduction problem. 
Many deep learning approaches \cite[e.g.][]{sutskever2014sequence,graves2012sequence} are not easily applicable to this task since we do not have a sufficiently large dataset of parallel texts.
Instead, we focus on approaches that mix hand-designed FSTs with trained language models, including deep learning approaches like RNNLMs \citep{mikolov2010}.

\subsection{End-to-end FSTs}
\label{sec:models:fsts}
Our initial approach represents the mapping from missionary to modern orthography using a composition of (weighted) FSTs; for a thorough review, see \citet{mohri1997finite}.

First, we construct a finite state acceptor, $I$, from the input text.
Here we construct a trivial chain-shaped acceptor that accepts only the input text. Each symbol in the input text is represented by a state which emits this symbol on a single transition that moves to the next state. The transition emitting the final symbol in the string leads to the sole accepting state.

\begin{figure*}[h]
  \centering
    \includegraphics[width=.9\linewidth]{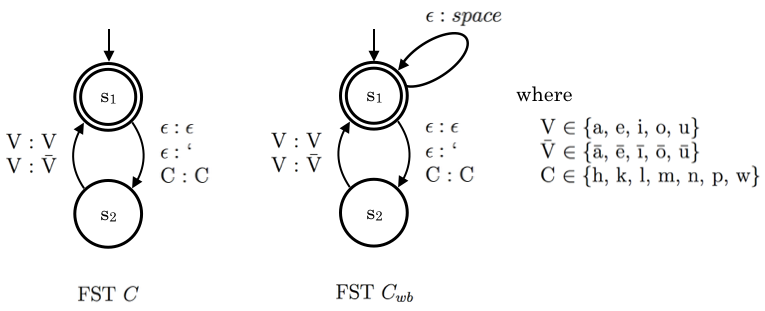}
  \caption{Two FSTs. The first, $C$, transduces between missionary and modern orthographies. The second, $C_{wb}$, introduces optional spaces (or word boundaries) after a vowel. In each FST, $s_1$ serves as both the initial and end state, while labelled arrows denote arcs. In the labelled transitions, V and \={V} are variables for short and long vowels, respectively, and C can be any consonant other than \orthog{`}. Some arcs, for upper-case letters, numbers, and punctuation, have been omitted for brevity.}
  \label{sec:fst-structure}
\end{figure*}

Second, we construct an FST called $C$, which models potential orthography changes that can occur when transliterating from the missionary to modern Hawaiian orthography (see \cref{sec:fst-structure}). For example, two non-deterministic transitions introduce an optional long-vowel map for \orthog{a}: $(\text{a} : \text{a})$ and $(\text{a} : \text{\=a})$. Another transition inserts glottal stops: $(\epsilon : \text{`})$. By capturing the orthographic changes we know to occur, the composition $I\circ C$ produces a large set of candidates to be narrowed using the language model. %

Third, we use the modern Hawaiian text from \cref{sec:data-preproc:corpus} to construct and evaluate a number of character-level n-gram language models, of various combinations of order and Katz backoff and Kneser-Ney (KN) smoothing \citep{katz1987estimation,kneser1995improved} (see \cref{sec:lm-training} for a list of models that we trained). 
N-gram language models can be expressed as weighted FSTs. 
We denote the n-gram or weighted FST language model as $G$. 
Character-level models are used as we wanted to generalize to out-of-vocabulary words, which we expect to occur frequently in a relatively small corpus like the one we have for Hawaiian. 

Finally, we use this model to infer modern orthography given a piece of text in missionary orthography as input,  then compose the FSTs to form the \emph{search graph} FST: $S = I\circ C\circ G$.  
The minimum cost path through $S$ gives the predicted modern orthography. %
Of these n-gram-based approaches, we found the Kneser-Ney-based models to perform best. 
These approaches will be referred to as \fstngram{$C$}{N}{KN}
and \fstngram{$C_{wb}$}{N}{KN}.

We circumvent the lack of a large, non-simulated parallel corpus by training the language model exclusively on text in the modern Hawaiian orthography. 
In turn, the orthographic transliteration FST $C$ produces candidates which are disambiguated by the language model.
The result is finally evaluated against the ground-truth modern text. 

Although the orthographic transliteration model is an approximation, and thus not exhaustive, %
it embodies an explicit and interpretable representation that can be extended independently of the rest of the model. %
To illustrate this, we constructed a variant $C_{wb}$ (where $wb$ stands for word boundary). 
$C_{wb}$ optionally inserts a space after each vowel using an additional arc that maps $(\epsilon : \textit{space})$ (again see \cref{sec:fst-structure}).
This variant is able to model some changes in Hawaiian's word-boundary conventions \citep{wilson1976}, such as \textit{alaila} becoming \textit{a laila} which demarcates the preposition \textit{a} `until' and noun \textit{laila} `then'. %
We report on the use of $C_{wb}$ to predict modern equivalents of 19th century newspaper samples in \cref{sec:results}.  
An example prediction can be found in \cref{sec:conclusions}, with more in \cref{sec:predictions}.

\subsection{Hybrid approach: Composing FSTs and RNNLMs}%
\label{sec:hybrid-models}
As an alternative approach, we tried combining the FST $C$ in the previous section with an RNNLM, since RNNLMs often generalize better than n-gram language models \citep{mikolov2010}.

An RNN %
is a neural network that models temporal or sequential data, by iterating a function mapping a state and input to a new state and output. %
These can be stacked to form a deep RNN. 
For language modelling, each step of the final RNN layer models a word or character sequence via $p(w_1,\dots,w_n) = \prod_{i=1}^n p(w_i | w_{1:i-1})$ and can be trained by maximum likelihood. 
Recent language modeling work has typically used the long short-term memory (LSTM) unit, 
because of its favorable gradient propagation properties \citep{hochreiter1997long}. 
All RNNs in this paper are LSTMs.

Our goal in the hybrid approach is to replace the n-gram language model in the end-to-end FST with an RNNLM.
While the minimum cost path through an FST
can be computed exactly as done in the previous section, it is not straightforward to compose the relation defined by an FST with an arbitrary one like that defined by an RNNLM.
Nonetheless, a minimum cost path through the composition of the FST and the RNNLM can be defined as a path (i.e.\ label sequence) that minimizes the sum of the FST path cost and the RNNLM cost. 

We can approximately find a minimum cost path for the composition of the two models by a breadth-first search over the FST graph, using a beam search, as follows. 
At any particular iteration, consider a single beam element.
The beam element holds the current FST and RNN states, and the path taken through the FST so far. 
We follow each possible arc from the current FST state, each producing a new child beam element, and feed the output symbol into the RNN (unless it is $\epsilon$). 
We note that there may be duplicate beam elements due the nondeterminicity of the FST, in which case the lower cost edge wins.
We sort by the sum of the FST and RNN costs, keep the lowest-cost $K$, and then proceed to the next iteration. 
If a beam element is on an accepting state of the FST, it is kept as-is between iterations.
Detailed pseudocode is provided in \cref{fig:alg}.

\begin{figure*}[htbp]
\begin{framed}
\begin{algorithmic}
\State $B \gets \{(\text{fst-state}=s_0, \text{rnn-state}=\mathbf{0}, \text{tokens}=[], \text{score}=0)\}$ \Comment{single initial beam element}
\While{$\exists b \in B : $ $b.\text{fst-state}$ is not final}
    \State $B' = \{\}$ \Comment{child beam elements}
    \For {\textbf{each} beam element $b \in B$ that is not in a final state}
        \For {\textbf{each} arc leaving $b.\text{fst-state}$ (new state $s'$, output $c$, weight $w$)}
            \State \Comment\parbox[t]{.8\linewidth}{
                Note that we do not need to use the input symbol for our purposes, but in some settings it may be useful to keep track of it.}
            \If{$b.\text{tokens}[-1] \neq \epsilon$}
                \State ($h'$, $p(y)$) = run RNN one timestep with input ($b.\text{rnn-state}$, $c$)
                \State rnncost = $-\log p(b.\text{tokens}[-1])$
                \State \Comment{denotes the negative log probability of the RNN's discrete distribution output at this timestep}
            \Else
                \State $h'$ = $b.\text{rnn-state}$
                \State rnncost = 0
            \EndIf
            \State $b' = \text{copy}(b)$
            \State $b'.\text{score} = b'.\text{score} + \text{rnncost} + w $
            \State $b'.\text{tokens}.\text{append}(c)$
            \State $b'.\text{rnn-state} = h'$
            \State $B' = B' \cup \{b'\}$
        \EndFor
    \EndFor
    \State $B' = B' \cup \{b\in B:b \text{ is in a final state}\}$
    \State Sort $B'$ by score, put best (lowest score) $K$ into $B$
\EndWhile
\end{algorithmic}
\end{framed}
\caption{Algorithm for approximately composing an FST with a RNN LM.}
\label{fig:alg}
\end{figure*}

Conceptually, this algorithm performs the same operation as the n-gram language model, except that we replace the n-gram language model with an RNN language model, and then we search over the FST graph, producing scores from FST weights and RNN's outputs.
Incidentally, we note that our implementation of the algorithm is slightly different than how we present it in the pseudocode, as we grouped RNN operations into batches for computational efficiency.

\section{Results}
\label{sec:results}

\subsection{Summary of language models trained}
\label{sec:lm-training}
In this paper, we will refer to the hybrid models as \fstrnn in general, and as \fstrnn-$C$ and \fstrnn-$C_{wb}$ if we want to distinguish which FST was used. 
The end-to-end FST models will similarly be referred to as FST-$C$ and FST-$C_{wb}$, with suffixes denoting what kind of n-gram and smoothing were used. 
For example, \fstngram{$C$}{7}{KN} denotes an end-to-end FST with a 7-gram language model and Kneser-Ney smoothing. 
Here is a list of n-gram language models that we considered (with perplexity scores in parentheses):
\begin{itemize}
\itemsep0em
\item 7-gram Katz backoff (3.15)
\item 7-gram Kneser-Ney  (3.07)
\item 7-gram Kneser-Ney with backoff  (3.15)
\item 9-gram Katz backoff (3.15)
\item \textbf{9-gram Kneser-Ney (2.95)} 
\item 9-gram Kneser-Ney with backoff (3.24)
\item 11-gram Katz backoff (3.39)
\item \textbf{11-gram Kneser-Ney (2.94)}
\item 11-gram Kneser-Ney with backoff (3.67)
\end{itemize}
We note that 13- and higher n-gram language models performed far worse. 
The numbers in parentheses show character-level perplexities produced using a validation set that was drawn from a synthetic parallel corpus, which will be explained in \cref{sec:sim-parallel-corpus}.

We also trained two character-level RNNLMs with the following configurations:
\begin{itemize}
\itemsep0em
\item 2 layers $\times$ 200 unit LSTM (2.70)
\item \textbf{3 layers $\times$ 200 unit LSTM (2.65)}
\end{itemize}
Both RNNLMs were trained with plain SGD, using a batch size of 30, a learning rate of 10, 
truncated backpropagation through time with 45 unrolling steps, and the gradient renormalized to norm 1 when it exceeded 1. 
Both RNNLMs used a dropout rate of 0.2 at the input, between RNN layers, and after the last RNN layer.
The best n-gram and RNNLMs are highlighted in \textbf{bold}. 
Overall, the best results were produced with the hybrid approach, using RNNLMs.

\subsection{Evaluation}
Because we were unable to find a sufficiently large corpus of parallel texts in the missionary and modern Hawaiian orthographies, we instead trained the n-gram and RNN language models on a corpus of modern Hawaiian texts (\emph{ground-truth}) (see \cref{sec:data-preproc:corpus}). %
Parallel corpora were only required to test predictions from missionary-era to modern texts, which were produced by composing one of the FSTs, $C$ or $C_{wb}$, with either an n-gram or RNN language model. 
To evaluate the accuracy of our approaches, we first derived a synthetic parallel corpus from our collection of modern Hawaiian texts. 
We also used a small but real parallel corpus, based on two 19th century newspaper texts and their hand-edited modern equivalents. 
Results based on these parallel corpora are reported in the following subsections.

\subsection{Simulated parallel corpus (modern texts)}
\label{sec:sim-parallel-corpus}
To produce a simulated parallel corpus (\emph{input-missionary}), we systematically reduced the orthography in the modern texts (\cref{sec:data-preproc:corpus}) using the backward mapping described above (\cref{sec:orthographies}). 
We then applied the end-to-end FST and hybrid FST-RNNLM models (\cref{sec:models}), with the aim of learning a forward mapping between orthographies that recovers the lost information.

We evaluated the predicted modern text (\emph{predictions}) by computing
\begin{equation*}
\text{CERR}=\frac{d(\text{prediction}, \text{ground-truth})}{d(\text{input-missionary}, \text{ground-truth})},
\end{equation*}
where $d$ denotes character-level edit distance. 
This is a modification of character error rate, normalized by the distance of the input and target rather than by the length of the target. 
We note that CERR may be high even when the predictions are very accurate as $d(\text{input-missionary}, \text{ground-truth})$ is small when the text is similar in both orthographies.

\cref{fig:result-table} gives more details about the strongest models from both  approaches. 
Out of the Kneser-Ney n-gram models, 
we found that the \fstngram{$C$}{9}{KN} and the version modelling word boundaries (\fstngram{$C_{wb}$}{9}{KN}) %
performed best on the synthetic parallel corpus and newspapers, respectively.
$C_{wb}$ was not applied to the synthetic parallel corpus as the synthetic parallel corpus did not model word splitting.
However, the hybrid models (\fstrnn) outperformed all end-to-end FSTs.

\begin{table*}[ht]
\centering
    \begin{tabular}{r|rr|r|rr}
    \toprule
    & \multicolumn{2}{c|}{LM perplexity} & \multicolumn{3}{c}{Transliteration performance (\%CERR)} \\
    Transliteration model
    & Valid. & Test & Corpus          & Newspaper 1           & Newspaper 2           \\
    \midrule
    \fstngram{($C$/$C_{wb}$)}{9}{KN}
    & 2.95 & 3.02 & 26.6\% & 50.7\% / 39.3\% & 52.5\% / 47.2\% \\
    \fstngram{($C$/$C_{wb}$)}{11}{KN}
    & 2.94 & 3.02 & 27.8\% & 53.9\% / 41.3\% & 54.1\% / 48.7\% \\
    \midrule
    \fstrnn-($C$/$C_{wb}$)
    &  \textbf{2.65}  & \textbf{2.69}   & \textbf{16.3\%} & 47.2\% / \textbf{34.3\%} & 49.8\% / \textbf{41.2\%} \\
    \bottomrule
    \end{tabular}
  \caption{Performance (\%CERR). Slash-separated pairs denote FSTs incapable/capable of inserting word boundaries, respectively; see \cref{sec:models}. The -KN suffix denotes Kneser-Ney smoothing.
  The data from \cref{sec:data-preproc:corpus} is used for evaluating the modern-orthography language model perplexity, and ``Corpus'' evaluates test-set transliteration performance from the synthetic missionary text back to the original modern text.}
  \vspace{-.5em}
  \label{fig:result-table}
\end{table*}

\subsection{Real parallel corpus (newspaper texts)}
Not content to evaluate the model on simulated missionary orthography, we also evaluated it on two newspaper texts, %
using selections originally published in 1867 and 1894 for which we had 19th century and manually-edited modern equivalents. 
The newspaper selections discuss \textit{Kahahana}, one of the last kings of O`ahu \citep{kamakau+2002}, and \textit{Uluhaimalama}, a garden party and secret political gathering, held after the deposition of Hawai`i's last queen \citep{pukui+2006}. 
Unlike the synthetic missionary corpus evaluation where we did not model word splitting, we found that replacing $C$ with $C_{wb}$ on the newspaper texts significantly improved the output, especially on the \fstrnn{} model.
Again, we found a hybrid model (\fstrnn-$C_{wb}$) to be the best performing model overall (\cref{fig:result-table}).

\section{Conclusions and future work}
\label{sec:conclusions}

With this paper we introduced a new transliteration problem to the field, that of mapping between old and new Hawaiian orthographies---where the modern Hawaiian orthography represents linguistic information that is missing from older missionary-era texts. 
One difficulty of this problem is that there is a limited amount of Hawaiian data, making data-hungry solutions like end-to-end deep learning unlikely to work. 
To solve the transliteration problem, we therefore proposed two models: the first was implemented end-to-end using weighted FSTs; the second was a hybrid deep learning approach that combined an FST and an RNNLM. 
Both models gave promising results, but the hybrid approach performed best. 
It allowed us to use a more powerful recurrent neural network-based language model, despite our dataset's small size. 
Factoring a problem like ours into one part that can be modelled exactly using expert domain knowledge and into another part that can be learned directly from data using deep learning is not novel; however it is a promising research direction for data-efficient modelling.
To our knowledge, this paper is the first to describe a procedure to compose an FST with an RNN by approximately performing beam search over the FST. %

While the role of the RNNLM part of the hybrid approach may be obvious, the FST component plays an important role too. 
For example, the hand-designed FST can be replaced without needing to re-train the RNNLM. 
We tried to showcase this modularity by constructing two FSTs which we referred to as $C$ and $C_{wb}$, where only the latter allowed the insertion of spaces. 
Future work could extend the FST to model orthographic changes suggested by an error analysis of the current model's predictions. 

\begin{table*}[h]
\begin{center}
  \begin{tabular}{r|l}
    \toprule
    \samp{Input} &Weheia ka Malapua Alii a Kanuia na Uluwehi no ia Wao.\\
    \samp{Prediction} &Wehe\correct{\textvisiblespace`}ia ka M\correct{\=a}la\correct{\textvisiblespace}\missing{p}ua Ali\correct{`}i a Kanu\missing{\textvisiblespace`}ia n\missing{a} Uluwehi n\wrong{\=o} ia Wao.\\
    \samp{Ground-truth} &Wehe `ia ka M\=ala Pua Ali`i a Kanu `ia n\=a Uluwehi no ia Wao.\\
    \bottomrule
  \end{tabular}
  \caption{Sample prediction. Data from from a 19th century Hawaiian newspaper \citep{pukui+2006}. Correct predictions are \correctmarker{}. 
  Characters omitted by the model as compared to the ground-truth are denoted by \missing{blue italics}; erroneous insertions or substitutions are denoted by \wrongmarker{}. To make white spaces explicit, we represent them with the symbol `\textvisiblespace'.
  More sample predictions can be found in \cref{sec:predictions}.
  }
  \vspace{-.5em}
  \label{fig:samples}
\end{center}
\end{table*}

An example of the current model's predictions (i.e.\ missionary input, predicted modern text, modern ground-truth) is given in \cref{fig:samples}. 
In this example, we see the model correctly predicting some word boundaries, glottal stops and long vowels; however, we note that the model could not predict uppercase \textit{Pua} (correct), because the input text contained lowercase \textit{pua} (incorrect), and no (p : P) transitions were included in $C$ or $C_{wb}$. 
Similar observations (see \cref{sec:predictions}) motivate new mappings for consonant substitutions like (r : l) and (s : k) that occur in loanword adaptations (e.g.~\textit{rose} $\Rightarrow$ \textit{loke}). 
The error analysis also motivates mappings to delete spaces $($\textvisiblespace~$ : \epsilon)$ and to handle contractions, like \textit{na'lii} $\Rightarrow$ \textit{n\={a} ali`i}. 
We could further incorporate linguistic knowledge of Hawaiian into the FST, which tells us, for example, about expected sequences of vowels \citep{parkerjones2010}. 
Additional improvements to the hybrid model might be obtained by increasing the amount of modern Hawaiian text used to train the RNNLM. 
One way to do this would be to accelerate the rate at which missionary-era Hawaiian texts are modernized. 
To this end, we hope that the present models will be used within the Hawaiian community to semi-automate, and thereby accelerate, the modernization of old Hawaiian texts.

\section*{Acknowledgments}
Mahalo to M.~Puakea Nogelmeier for providing an electronic copy of \textit{Hi`iakaikapoliopele} \citep{hooulumahiehie2007}. 
We gratefully acknowledge the support of the NVIDIA Corporation for donating the Titan X(p) GPUs used in this research.
We also gratefully acknowledge the ACL for allowing us to reproduce parts of a shorter (5 page) version of this work \citep{shillingford+2018}.

\bibliography{refs}
\bibliographystyle{acl_natbib_nourl}

\clearpage

\onecolumn

\appendix

\section{Appendix}

\subsection*{Sample predictions from newspaper data}
\label{sec:predictions}
Each block of three lines begins with the input, followed by the prediction, followed by the ground-truth. %
We denote characters omitted by the model as compared to the ground-truth by \missingmarker{}, whereas characters that are erroneously inserted or substituted (or should have been substituted) for another character are marked with \wrongmarker{}. When the incorrect character is a space, the space is replaced with `\textvisiblespace'.
Note that correct changes are left unmarked.

\subsubsection*{First 10 sentences in Newspaper 1:}
\samp{Input 1} Weheia ka Malapua Alii a Kanuia na Uluwehi no ia Wao.
\samp{Prediction 1} Wehe `ia ka M\=ala \wrong{p}ua Ali`i a Kanu\missing{\textvisiblespace`}ia n\wrong{a} Uluwehi n\wrong{\=o} ia Wao.
\samp{Ground-truth 1} Wehe `ia ka M\=ala Pua Ali`i a Kanu `ia n\=a Uluwehi no ia Wao.

\samprule

\samp{Input 2} E like no hoi me ka mea i hoike akea ia ae no ka manawa a me ka la e weheia ai a e kanuia ai hoi o na pua a me na mea ulu e ae ma kahi nona ka inoa kilakila maluna ae, pela no i hoea io mai ai i ka Poaha iho la, hora 9 A. M. a mahope mai. Mamua ae o ia manawa, ua lehulehu na poe i pii aku me na mea kanu, maluna o na kaa a malalo no hoi.
\samp{Prediction 2} E like n\=o ho`i me ka mea i h\=o`ike \=akea `ia a`e no ka manawa a me ka l\=a e wehe `ia ai a e kanu `ia ai ho`i \wrong{`}o n\=a pua a me n\=a mea ulu `\=e a`e ma kahi nona ka inoa kilakila ma luna a`e, p\=el\=a n\=o i h\=o`ea `i`o mai ai i ka P\wrong{o}`ah\=a iho\wrong{ }la, ho\wrong{r}a 9 A. M. a ma hope mai. Ma mua a`e o ia manawa, ua lehulehu n\=a po`e i pi`i aku me n\=a mea kanu, ma luna o n\=a ka`a a ma lalo n\=o ho`i.
\samp{Ground-truth 2} E like n\=o ho`i me ka mea i h\=o`ike \=akea `ia a`e no ka manawa a me ka l\=a e wehe `ia ai a e kanu `ia ai ho`i o n\=a pua a me n\=a mea ulu `\=e a`e ma kahi nona ka inoa kilakila ma luna a`e, p\=el\=a n\=o i h\=o`ea `i`o mai ai i ka P\=o`ah\=a ihola, hola 9 A. M. a ma hope mai. Ma mua a`e o ia manawa, ua lehulehu n\=a po`e i pi`i aku me n\=a mea kanu, ma luna o n\=a ka`a a ma lalo n\=o ho`i.

\samprule

\samp{Input 3} Hoomakaia ke Kanu Ana.
\samp{Prediction 3} Ho`omaka `ia ke Kanu `Ana.
\samp{Ground-truth 3} Ho`omaka `ia ke Kanu `Ana.

\samprule

\samp{Input 4} Ua hoea ae no ilaila ka Puali Puhiohe Lahui, a i ka aneane ana ae i ka manawa, a i ole ia, ua hala no paha he hapalua hora mahope iho o ka hora 9, ua uhene mai la lakou i ke mele Liliuokalani, a o ka wa no ia o Kamalii Kawananakoa, ma ka aoao o ke Aliiaimoku, i kanu iho ai i kekahi kumu lehua o Mokaulele iwaenakonu, i hoopuniia ae me na ohawai a me kekahi mau mea kanu Hawaii e ae iloko o kekahi ponaha poepoe, a makai iho hoi o ia wahi i kanu ai o Kamalii Kalanianaole i kekahi kumu lehua ahihi ma ka aoao o ke Alii ka Moiwahine Kanemake. Pau keia mau hana ae la, ua noa i na mea a pau, a ua hele no hoi ia wahi a eeu i na oiwi palupalu o kakou, e kanu ana i kela a me keia mea, a he mau oiwi oolea no hoi kekahi malaila e kokua ana. He mea makehewa paha ke helupapa aku i na pua i kanuia. Hookahi a makou mea i mahalo, oia no kekahi wahi i kanu mua e ia no makai iki mai o ka puka komo, me ka inoa o ia kihapai e kau ae la maluna, a he ku maoli no i ka nani.
\samp{Prediction 4} Ua h\=o`ea a`e n\=o i laila ka \wrong{P}\=u`ali \wrong{P}uhi `ohe \wrong{L}\=ahui, a i ka `ane`ane `ana a`e i ka manawa, a i `ole ia, ua hala n\=o paha he hapalua ho\wrong{r}a ma hope iho o ka ho\wrong{r}a 9, ua `uhene mai\wrong{ }la l\=akou i ke mele Lili`uokalani, a `o ka w\=a n\=o ia o \wrong{K\=a}m\wrong{a}li`i Kaw\=ananakoa, ma ka `ao`ao o ke \wrong{A}li`i `ai moku, i kanu iho ai i kekahi kumu lehua o M\wrong{\=o}kaulele i waenakonu, i ho`opuni `ia a`e me n\=a `\wrong{o}h\=a\wrong{ }wai a me kekahi mau mea kanu Hawai`i `\=e a`e i loko o kekahi p\=onaha poepoe, a ma kai iho ho`i o ia wahi i kanu ai `o \wrong{K}am\=ali`i Kalaniana`ole i kekahi kumu lehua `\=ahihi ma ka `ao`ao o ke \wrong{A}li`i\missing{,} ka \wrong{M}\=o`\=\i{}\missing{\textvisiblespace}wahine \wrong{K}\=ane make. Pau k\=eia mau hana a`e\wrong{ }l\wrong{\=a}, ua noa i n\=a mea a pau, a ua hele n\=o ho`i ia wahi a \missing{`}e\wrong{ }`eu i n\=a `\=oiwi palupalu o k\=akou, e kanu ana i k\=el\=a a me k\=eia mea, a he mau `\=oiwi `o`ole`a n\=o ho`i kekahi ma laila e k\=okua ana. He mea makehewa paha ke helu papa aku i n\=a pua i kanu `ia. Ho`okahi a m\=akou mea i mahalo, `o ia n\=o kekahi wahi i kanu mua \wrong{`\=e} \wrong{`}ia n\=o ma kai iki mai o ka puka komo, me ka inoa o ia k\=\i{}h\=apai e kau a`e\wrong{ }l\wrong{\=a} ma luna, a he k\=u maoli n\=o i ka nani.
\samp{Ground-truth 4} Ua h\=o`ea a`e n\=o i laila ka p\=u`ali puhi `ohe l\=ahui, a i ka `ane`ane `ana a`e i ka manawa, a i `ole ia, ua hala n\=o paha he hapalua hola ma hope iho o ka hola 9, ua `uhene maila l\=akou i ke mele Lili`uokalani, a `o ka w\=a n\=o ia o kam\=ali`i Kaw\=ananakoa, ma ka `ao`ao o ke ali`i `ai moku, i kanu iho ai i kekahi kumu lehua o Mokaulele i waenakonu, i ho`opuni `ia a`e me n\=a `\=oh\=awai a me kekahi mau mea kanu Hawai`i `\=e a`e i loko o kekahi p\=onaha poepoe, a ma kai iho ho`i o ia wahi i kanu ai `o kam\=ali`i Kalaniana`ole i kekahi kumu lehua `\=ahihi ma ka `ao`ao o ke ali`i, ka m\=o`\=\i{} wahine k\=ane make. Pau k\=eia mau hana a`ela, ua noa i n\=a mea a pau, a ua hele n\=o ho`i ia wahi a `e`eu i n\=a `\=oiwi palupalu o k\=akou, e kanu ana i k\=el\=a a me k\=eia mea, a he mau `\=oiwi `o`ole`a n\=o ho`i kekahi ma laila e k\=okua ana. He mea makehewa paha ke helu papa aku i n\=a pua i kanu `ia. Ho`okahi a m\=akou mea i mahalo, `o ia n\=o kekahi wahi i kanu mua e ia n\=o ma kai iki mai o ka puka komo, me ka inoa o ia k\=\i{}h\=apai e kau a`ela ma luna, a he k\=u maoli n\=o i ka nani.

\samprule

\samp{Input 5} Ka Lanai Pea Ahaaina.
\samp{Prediction 5} Ka L\=ana\wrong{`}i Pe`a `Aha`aina.
\samp{Ground-truth 5} Ka L\=anai Pe`a `Aha`aina.

\samprule

\samp{Input 6} Mauka iki aku hoi o keia mala, ma ka aoao maluna o ka hale noho o J. Mana, ua kukuluia ae la he lanai pea, a ilaila kahi i hoomakaukauia ai o kekahi papaaina. E houhene mau ana no hoi ka puali puhiohe i kela a me keia wa a hiki i ka wa ai, i ke kau pono o ka la i ka piko a mahope iho paha, me ka hoomau aku no i ka hoolealea ana a pau ka papaaina mua a ke Alii, o na Kamaliikane no ma kona a me ko kona kaikoeke wahi. Ekolu papaaina i hoonohoia ai a ua ai na poe a pau i hiki aku a lawa pono, me ka hoounaia o kahi mea-ai no kekahi poe i maopopo i hiki ole aku.
\samp{Prediction 6} Ma uka iki aku ho\wrong{`}i o k\=eia m\=ala, ma ka `ao`ao ma luna o ka hale noho o J. M\wrong{\=a}n\wrong{\=a}, ua k\=ukulu `ia a`e\wrong{ }la he l\=anai pe`a, a i laila kahi i ho`om\=akaukau `ia ai \wrong{`}o kekahi papa\missing{\textvisiblespace}`\wrong{a}ina. E h\wrong{o}`uhene mau ana n\=o ho`i ka p\=u`ali puhi `ohe i k\=el\=a a me k\=eia w\=a a hiki i ka w\=a `ai, i ke kau pono o ka l\=a i ka piko a ma hope iho paha, me ka ho`omau aku n\=o i ka ho`ole`ale`a `ana a pau ka papa `aina mua a ke \wrong{A}li`i, \wrong{`}o n\=a \wrong{K}am\=ali`i\missing{\textvisiblespace}k\=ane n\=o ma k\wrong{o}na a me k\wrong{o} k\wrong{o}na kaiko`eke wahi. `Ekolu papa `\wrong{\=a}ina i ho`onoho `ia ai a ua `ai n\=a po`e a pau i hiki aku a lawa pono, me ka ho`ouna `ia o kahi mea\wrong{ -}`ai no kekahi po`e i maopopo i hiki `ole aku.
\samp{Ground-truth 6} Ma uka iki aku hoi o k\=eia m\=ala, ma ka `ao`ao ma luna o ka hale noho o J. Mana, ua k\=ukulu `ia a`ela he l\=anai pe`a, a i laila kahi i ho`om\=akaukau `ia ai o kekahi papa `\=aina. E h\=o`uhene mau ana n\=o ho`i ka p\=u`ali puhi `ohe i k\=el\=a a me k\=eia w\=a a hiki i ka w\=a `ai, i ke kau pono o ka l\=a i ka piko a ma hope iho paha, me ka ho`omau aku n\=o i ka ho`ole`ale`a `ana a pau ka papa `aina mua a ke ali`i, o n\=a kam\=ali`i k\=ane n\=o ma k\=ona a me k\=o k\=ona kaiko`eke wahi. `Ekolu papa `aina i ho`onoho `ia ai a ua `ai n\=a po`e a pau i hiki aku a lawa pono, me ka ho`ouna `ia o kahi mea`ai no kekahi po`e i maopopo i hiki `ole aku.

\samprule

\samp{Input 7} O na keiki puhiohe no hoi kekahi i kanu mau wahi mea kanu, a he mau mea ulu Hawaii wale no hoi ka lakou o ke ano papa kahuna. Mamuli o ka oluolu a me ka lokomaikai o kahi o lakou, ua loaa mai ia makou ka papa hoike a me na wehewehe ana o ka lakou.
\samp{Prediction 7} `O n\=a keiki puhi `ohe n\=o ho`i kekahi i ka\wrong{ }n\wrong{\=u} mau wahi mea kanu, a he mau mea ulu Hawai`i wale n\=o ho`i k\wrong{\=a} l\=akou o ke `ano papa kahuna. Ma muli o ka `olu`olu a me ka lokomaika`i o k\missing{ek}ahi o l\=akou, ua loa`a mai i\=a m\=akou ka papa h\=o`ike a me n\=a wehewehe `ana o k\=a l\=akou.
\samp{Ground-truth 7} `O n\=a keiki puhi `ohe n\=o ho`i kekahi i kanu mau wahi mea kanu, a he mau mea ulu Hawai`i wale n\=o ho`i ka l\=akou o ke `ano papa kahuna. Ma muli o ka `olu`olu a me ka lokomaika`i o kekahi o l\=akou, ua loa`a mai i\=a m\=akou ka papa h\=o`ike a me n\=a wehewehe `ana o k\=a l\=akou.

\samprule

\samp{Input 8} Mau Mea Kanu Hookalakupua
\samp{Prediction 8} Mau Mea Kanu Ho`okalakupua\missing{.}
\samp{Ground-truth 8} Mau Mea Kanu Ho`okalakupua.

\samprule

\samp{Input 9} o ke ano hookahuna oiaio maoli no. Eia iho no ia papa hoike:
\samp{Prediction 9} `\wrong{o} ke `ano ho`okahuna `oia`i`o maoli n\=o. Eia iho n\=o ia papa h\=o`ike:
\samp{Ground-truth 9} `O ke `ano ho`okahuna `oia`i`o maoli n\=o. Eia iho n\=o ia papa h\=o`ike:

\samprule

\samp{Input 10} hala polapola. Ko puni e Kalani o ka lei e, leiia hoi o Halaomapuana, onaona i ka ihu, huihui ke hanu iho.
\samp{Prediction 10} hala polapola. K\=o puni \wrong{\=e} Ka\wrong{ }lani o ka lei \=e, lei `ia ho`i `o H\wrong{\=a}laom\wrong{a}puana, onaona i ka ihu, hu`ihu`i ke hanu iho.
\samp{Ground-truth 10} hala polapola. K\=o puni e Kalani o ka lei \=e, lei `ia ho`i `o Halaom\=apuana, onaona i ka ihu, hu`ihu`i ke hanu iho.

\subsubsection*{First 10 sentences in Newspaper 2:}
\samp{Input 1} O Kahahana, he alii kapu ia no o Oahu.
\samp{Prediction 1} `O Kahahana, he ali`i kapu ia n\wrong{\=o }o O`ahu.
\samp{Ground-truth 1} `O Kahahana, he ali`i kapu ia no O`ahu.

\samprule

\samp{Input 2} O Kaionuilalahai ka makuahine, ka moopuna a Kalaniomaiheuila, ke kaikamahine a Kalanikahimakaialii, a laua o Kualu ke kaikuahine, a mua hoi o Kaulahea ka Moi o Maui.
\samp{Prediction 2} `O Ka\missing{`}i\wrong{`}onuilalahai ka makuahine, ka mo`opuna a Kalani\missing{`}\wrong{ o }ma\wrong{`}i\wrong{ }he\wrong{ }uila, ke kaikamahine a Kalani\wrong{ }kahi\wrong{ }maka\missing{`}\wrong{ }i\wrong{ }ali`i, \wrong{`\=a} l\=aua `o K\=u\wrong{`}alu ke kaikuahine, a mua ho`i o Ka`ula\wrong{ }hea\missing{,} ka M\=o`\=\i{} o Maui.
\samp{Ground-truth 2} `O Ka`ionuilalahai ka makuahine, ka mo`opuna a Kalani`\=omaiheuila, ke kaikamahine a Kalanikahimaka`iali`i, a l\=aua `o K\=ualu ke kaikuahine, a mua ho`i o Ka`ulahea, ka M\=o`\=\i{} o Maui.

\samprule

\samp{Input 3} A o ka makuakane, oia hoi o Elani, no ka ohana a Kupanihi, a o Keopuolani.
\samp{Prediction 3} A `o ka makua k\=ane, `o ia ho`i `o \missing{`}\wrong{E}lani, no ka `ohana a K\wrong{u}p\wrong{a}nihi, a `o Ke\=op\=uolani.
\samp{Ground-truth 3} A `o ka makua k\=ane, `o ia ho`i `o `\=Elani, no ka `ohana a K\=up\=anihi, a `o Ke\=op\=uolani.

\samprule

\samp{Input 4} I ko Kahahana manawa kamalii, ua kii ia mai e Kahekili e lawe i Maui i keiki nana.
\samp{Prediction 4} I ko Kahahana manawa kamali`i, ua ki`i `ia mai e Kahekili e lawe i Maui i keiki n\=ana.
\samp{Ground-truth 4} I ko Kahahana manawa kamali`i, ua ki`i `ia mai e Kahekili e lawe i Maui i keiki n\=ana.

\samprule

\samp{Input 5} Ua hanai kapu ia oia i Maui.
\samp{Prediction 5} Ua h\=anai kapu `ia `o ia i Maui.
\samp{Ground-truth 5} Ua h\=anai kapu `ia `o ia i Maui.

\samprule

\samp{Input 6} A i kona lilo ana ae i kanaka makua, ua lilo oia i kanaka maikai, a ua nani hoi kona helehelena, a ua piipii maikai kona lauoho; a ua kapaia oia i kekahi wa, he piipii hahai moa.
\samp{Prediction 6} A i kona lilo `ana a`e i kanaka m\wrong{\=a}kua, ua lilo `o ia i kanaka maika`i, a ua nani ho`i kona helehelena, a ua pi`ipi`i maika`i kona lauoho; a ua kapa `ia `o ia i kekahi w\=a, he pi`ipi`i hahai moa.
\samp{Ground-truth 6} A i kona lilo `ana a`e i kanaka makua, ua lilo `o ia i kanaka maika`i, a ua nani ho`i kona helehelena, a ua pi`ipi`i maika`i kona lauoho; a ua kapa `ia `o ia i kekahi w\=a, he pi`ipi`i hahai moa.

\samprule

\samp{Input 7} I kona lilo ana i kanaka makua, hooipoipo aku la oia me na wahine kaukaualii, nolaila, ua ali ia ke kapu, a ua kapa hou ia ka inoa o Walia, a ua hoopauia ka inoa Ahi, Wela, Hahana.
\samp{Prediction 7} I kona lilo `ana i kanaka m\wrong{\=a}kua, ho`oipoipo aku\wrong{ }la `o ia me n\=a w\=ahine kaukauali`i, no laila, ua `ali `ia ke kapu, a ua kapa hou `ia ka inoa `o Wali\wrong{`}a, a ua ho`opau `ia ka \wrong{`}ino\wrong{ }a \wrong{`}Ahi, Wela, Hahana.
\samp{Ground-truth 7} I kona lilo `ana i kanaka makua, ho`oipoipo akula `o ia me n\=a w\=ahine kaukauali`i, no laila, ua `ali `ia ke kapu, a ua kapa hou `ia ka inoa `o Walia, a ua ho`opau `ia ka inoa Ahi, Wela, Hahana.

\samprule

\samp{Input 8} Lawe ae la o Kahahana i wahine nana, o Kekuapoi ka inoa.
\samp{Prediction 8} Lawe a`e\wrong{ }la `o Kahahana i wahine n\=ana,\wrong{ }`o Kekuapo`i ka inoa.
\samp{Ground-truth 8} Lawe a`ela `o Kahahana i wahine n\=ana,`o Kekuapo`i ka inoa.

\samprule

\samp{Input 9} Ua olelo ia no hoi kela wahine, aohe ona lua iloko o ke aupuni Hawaii nei, a ua kaulana kona inoa mai Hawaii a Kauai, a ua lilo hoi i kaao, a ua kiekie hanohano kona kino; aole no hoi i ike ia kahi kina mai luna o ke poo a hiki i na kapuai wawae; he mau maka manu nunu kona i like me ko ka mohoea; a ua like hoi na helehelena o na hiohiona papalina me ka opuu rose i mohala maikai i ke kakahiaka; a ua nui na loli o kona mau hiohiona, i ke kakahiaka, i ke awakea, ahiahi a me ka po; o ka puo kelakela, oiai oia iloko o ia manawa, ua kupono ke ali ana o ke kapu.
\samp{Prediction 9} Ua `\=olelo `ia n\=o ho`i k\=el\=a wahine, `a`ohe ona lua i loko o ke aupuni Hawai`i nei, a ua kaulana kona inoa mai Hawai`i a Kaua`i, a ua lilo ho`i i ka`ao, a ua ki`eki`e hanohano kona kino; `a`ole n\=o ho`i i `ike `ia kahi \wrong{`}k\wrong{i}n\wrong{a} mai luna o ke po`o a hiki i n\=a kapua`i w\=awae; he mau maka manu n\=un\=u kona i like me ko ka moho\wrong{ }e\wrong{ `\=a}; a ua like ho`i n\=a helehelena o n\=a hi`ohi`ona p\=ap\=alina me ka `\=opu`u \wrong{r}o\wrong{s}e i m\wrong{o}hala maika`i i ke kakahiaka; a ua nui n\=a loli o kona mau hi`ohi`ona, i ke kakahiaka, i ke awakea, ahiahi a me ka p\=o; `o ka pu\missing{`}\wrong{o} kelakela, `oiai `o ia i loko o ia manawa, ua k\=upono ke `ali `ana o ke kapu.
\samp{Ground-truth 9} Ua `\=olelo `ia n\=o ho`i k\=el\=a wahine, `a`ohe ona lua i loko o ke aupuni Hawai`i nei, a ua kaulana kona inoa mai Hawai`i a Kaua`i, a ua lilo ho`i i ka`ao, a ua ki`eki`e hanohano kona kino; `a`ole n\=o ho`i i `ike `ia kahi k\=\i{}n\=a mai luna o ke po`o a hiki i n\=a kapua`i w\=awae; he mau maka manu n\=un\=u kona i like me ko ka mohoea; a ua like ho`i n\=a helehelena o n\=a hi`ohi`ona p\=ap\=alina me ka `\=opu`u loke i m\=ohala maika`i i ke kakahiaka; a ua nui n\=a loli o kona mau hi`ohi`ona, i ke kakahiaka, i ke awakea, ahiahi a me ka p\=o; `o ka pu`\=o kelakela, `oiai `o ia i loko o ia manawa, ua k\=upono ke `ali `ana o ke kapu.

\samprule

\samp{Input 10} Ua uluhua na'lii, na kahuna a me na makaainana o ke aupuni o Oahu i ko lakou Moi ia Kumahana, i ke keiki a Peleioholani.
\samp{Prediction 10} Ua uluhua n\=a \wrong{`'}\missing{a}li\missing{`}i, n\=a k\=ahuna a me n\=a maka`\=ainana o ke aupuni o O`ahu i ko l\=akou M\=o`\=\i{}\missing{,} i\=a K\=umahana, i ke keiki a Pelei\=oh\=olani.
\samp{Ground-truth 10} Ua uluhua n\=a ali`i, n\=a k\=ahuna a me n\=a maka`\=ainana o ke aupuni o O`ahu i ko l\=akou M\=o`\=\i{}, i\=a K\=umahana, i ke keiki a Pelei\=oh\=olani.

\end{document}